\documentclass{article}
\usepackage{spconf,amsmath,graphicx}
\usepackage{pgfplots}
\usepackage{subcaption}
\usepackage{booktabs}
\usepackage[hidelinks]{hyperref}

\title{Mixture-of-Experts-based Entropy Model for\\ Learned Image Compression}
\name{Jonas Brenig,\quad Radu Timofte\thanks{This work was supported by the Alexander von Humboldt Foundation.}}
\address{Computer Vision Lab, CAIDAS \& IFI, University of W\"urzburg, Germany}

\usetikzlibrary{positioning,arrows,bending,fit,backgrounds}
\pgfplotsset{
  vvc/.style={
    color=black,
    mark=none,
  },
  tcm/.style={
    color=blue,
    mark=none %
    mark options={scale=1, fill=blue!50!white},
    thick,
  },
  mlicpp/.style={ 
    color=orange,
    mark=none %
    mark options={scale=1, fill=orange!50!white},
    thick,
  },
  lalic/.style={
    color=magenta,
    mark=none %
    mark options={scale=1, fill=purple!50!white},
    thick,
  },
  dcae/.style={
    color=red,
    mark=none %
    mark options={scale=1, fill=red!50!white},
    thick,
  },
  elic/.style={
    color=purple,
    mark=none %
    mark options={scale=1, fill=purple!50!white},
    thick,
  },
  cca/.style={
    color=brown,
    mark=none %
    mark options={scale=1, fill=brown!50!white},
    thick,
  },
  moe/.style={
    color=cyan,
    mark=*,
    mark options={scale=1, fill=cyan!50!white},
    thick,
    line width=1.4pt
  },
}
  
\begin{document}
\maketitle
\begin{abstract}
Learned image compression has seen significant progress in recent years with the development of end-to-end learned models %
that achieve better compression efficiency than state-of-the-art conventional methods.
Recently, Mixture of Experts (MoE) approaches have seen promising results in NLP and computer vision tasks.
In this paper, we introduce the MoE approach to learned image compression. 
We propose a MoE-based Entropy model (MoEE) for learned image compression, allowing the model to selectively activate only the subset of parameters required for the input image.
Our model achieves a BD-Rate improvement over VVC of -16.85\% on the Kodak dataset.
\end{abstract}
\begin{keywords}
mixture of experts, learned image compression
\end{keywords}
\section{Introduction}
\label{sec:intro}

With the vast number of images being captured and shared daily, efficient image compression techniques are essential to reduce storage requirements and transmission bandwidth. Traditional image compression methods, such as JPEG~\cite{wallace1991jpeg}, have been widely used for decades. 
Newer codecs, like AVIF and VVC~\cite{dominguez2022versatile}, have improved compression efficiency but still rely on hand-crafted algorithms.

Learned Image Compression (LIC) has emerged as a promising alternative, leveraging deep learning to optimize compression performance. LIC methods typically employ autoencoders to learn compact representations of images, achieving superior rate-distortion performance compared to traditional codecs~\cite{minnen2018joint,minnen2020channel,he2021checkerboard,he2022elic,jiang2023mlic,lu2025learned,brenig2026msl}. 
In order to better adapt to the diverse characteristics of images, the recently proposed DCAE~\cite{lu2025learned} employs a dictionary-based entropy model.

Recently, Mixture of Experts methods have been popularized in various fields such as NLP and various computer vision tasks~\cite{luo2023wm,riquelme2021scaling}. Mixture of Experts network designs promise to increase the number of parameters in the network without significantly increasing computational complexity. This allows the network to learn more complex representations while maintaining efficient inference.

\begin{figure}[th]
  \centering
  \begin{tikzpicture}
    \scriptsize
    \begin{axis}[
        yticklabel style={
            /pgf/number format/precision=2
        },
        xticklabel style={
            /pgf/number format/fixed,
            /pgf/number format/precision=3
        },
        scaled x ticks=false,
        xlabel=Decoding latency \lbrack ms\rbrack \textdownarrow,
        ylabel style = {align=center},
        ylabel=BD-Rate (Anchor: VTM-17.0) \lbrack \%\rbrack \textdownarrow,
        ylabel near ticks,
        width=\columnwidth,
        height=8cm,
        mark size=4pt,
        grid=both,
        grid style={dashdotted},
        scatter/classes={
            moe={cyan}, 
            dcae={pink},
            mlic={orange},
            lalic={magenta},
            elic={purple},
            tcm={blue},
            c2f={green},
            cca={brown}
        },
        y dir=reverse,
    ]        
    \addplot[
        scatter, 
        only marks,
        scatter src=explicit symbolic,
        nodes near coords*={\annotvalue },
        node near coord style={anchor=north, font=\scriptsize, yshift=\shiftvalue, xshift=\xshiftvalue},
        visualization depends on={value \thisrow{annotation} \as \annotvalue},
        visualization depends on={ value \thisrow{shift} \as \shiftvalue},
        visualization depends on={ value \thisrow{xshift} \as \xshiftvalue},
        ] 
        table[meta=label,x expr={\thisrow{x}*1000}] {
            x       y                       label  annotation shift xshift
            0.07298389077186584    -16.184562097550337     moe    {MoEE (ours)} +13pt 0
            0.06733209292093913   -15.065656860849174     dcae   {DCAE (CVPR'25)} -4pt -6pt
            0.09827747742335001   -14.687692122427476      mlic   {MLIC++ (ICMLW'23)} +13pt -17pt
            0.09698066711425782    -13.746354022421436      lalic  {LALIC (CVPR'25)} -3pt -8pt
            0.06119625568389893    -2.400909387090977      elic   {ELIC (CVPR'22)} -3pt 10pt
            0.09303197463353476    -5.434094735039919       tcm    {TCM (CVPR'23)} -3pt 0
            0.08299588759740194    -10.891394873146233     cca    {CCA (NeurIPS'24)} -3pt 0
        };
    \end{axis}
\end{tikzpicture}
  \caption{BD-Rate vs. Latency comparison on the Kodak dataset (Anchor VTM-17.0). Top-left is better. Our MoE-based entropy model achieves the best BD-Rate while maintaining competitive decoding latency.}
\end{figure}
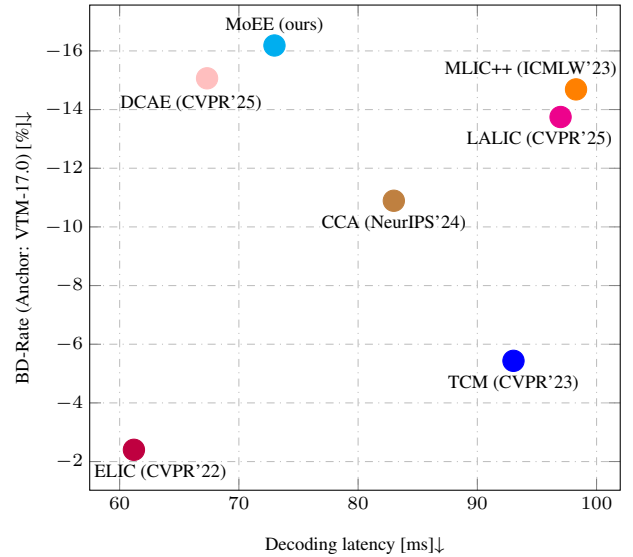

In this work, we propose a novel approach that incorporates a Mixture of Experts (MoE) framework into learned image compression. %
Building on the established grouped-channel entropy model~\cite{he2022elic,jiang2023mlic,lu2025learned,feng2025linear}, we introduce MoE-based context modeling (MoEE), which further enhances the entropy estimation accuracy. 
By combining the strengths of Mixture of Experts with learned image compression, our approach achieves improved performance while maintaining computational efficiency.

Our contributions are as follows:
\begin{itemize}
    \item We introduce a Mixture-of-Experts-based entropy model (MoEE) for learned image compression, increasing the capacity of the context prediction network, without strongly impacting the decoding speed.
    \item We conduct extensive experiments demonstrating that our MoE-based LIC model outperforms state-of-the-art learned and traditional image compression methods on standard benchmarks.
\end{itemize}

\section{Related Work}
\label{sec:related}

\subsection{Learned Image Compression}
Most recent work in learned image compression (LIC) is based on learned nonlinear transformations that are trained end-to-end on large datasets. These models typically consist of an encoder, a hyper-prior-based entropy model, and a decoder. 
The hyper-prior approach was first introduced by Ballé et al.~\cite{balle2018variational} and has since been extended to use autoregressive entropy modeling to improve compression efficiency~\cite{minnen2018joint,minnen2020channel}. 

However, fully autoregressive and channel-wise autoregressive models are computationally expensive and require significant computational resources and training time.
To address this, subsequent works have focused on improving compression efficiency by refining entropy models.
This includes using a checkerboard pattern in entropy modeling to capture spatial redundancies~\cite{he2021checkerboard} or using channel groups instead of decoding each channel separately~\cite{minnen2020channel,he2022elic}.
Ali et al.~\cite{ali2023towards} propose to replace autoregressive context modeling with a decorrelation loss to improve compression efficiency.
Other approaches focus on improving autoregressive entropy modeling by incorporating attention-based context prediction~\cite{jiang2023mlic++,jiang2025mlicv2,qian2022entroformer,zou2022devil} or using latents with multiple scales~\cite{brenig2026msl}.
The recently proposed DCAE~\cite{lu2025learned} introduces an entropy model based on dictionary learning. This method achieves state-of-the-art compression efficiency and outperforms previous methods on both compression efficiency and computational complexity. 

With the recent success of the transformer architecture, several works have explored the use of transformers in LIC~\cite{liu2023learned,zou2022devil,zhu2022transformer,qian2022entroformer,li2025learned}, often using the Swin-Transformer~\cite{liu2021swin} to replace the traditional CNN architecture. 
Recent works have tried to bring architectural improvements from other fields of deep learning, such as selective state spaces~\cite{gu2024mamba,qin2024mambavc} and RWKV blocks~\cite{peng2023rwkv,feng2025linear}, to the field of image compression.

\subsection{Mixture of Experts}
Mixture of Experts (MoE) architectures consist of multiple expert networks and a gating network that selects the most appropriate expert for a given input. 
This allows the model to leverage the strengths of each expert, leading to improved performance on a variety of tasks, while keeping the computational cost low by only activating a subset of experts for each input~\cite{shazeer2017outrageously}.

Mixture of Experts have been widely used in NLP tasks, and been recently introduced to a variety of computer vision tasks as well~\cite{riquelme2021scaling,puigcerver2024from,shazeer2017outrageously,xue2024openmoe}.

For low-level vision tasks, Luo et al.~\cite{luo2023wm} proposed a MoE-based architecture for weather artifact removal.
Zamfir et al.~\cite{zamfir2025complexity} proposed a complexity-aware Mixture of Experts architecture for all-in-one image restoration tasks.

\section{Method}
\label{sec:method}

\begin{figure*}[th]
  \centering
  \begin{subfigure}[t]{0.49\textwidth}
    \centering
    \usetikzlibrary{positioning,arrows,bending}
\begin{tikzpicture}[
    module/.style={draw, thick, rounded corners, minimum width=15ex, minimum height=4ex,align=center},
    embmodule/.style={module, fill=red!20},
    mhamodule/.style={module, fill=orange!20},
    lnmodule/.style={module, fill=yellow!20},
    ffnmodule/.style={module, fill=cyan!20},
    encoder/.style={module, fill=green!20, minimum width=15ex},
    decoder/.style={module, fill=green!20, minimum width=15ex},
    entropy/.style={module, fill=blue!20, minimum width=15ex,text width=15ex,},
    projector/.style={module, fill=green!20, minimum width=7ex},
    arrow/.style={-stealth', thick, rounded corners},
    connector/.style={thick, rounded corners},
    coder/.style={module, minimum width=5ex, fill=yellow!20},
    helper/.style={inner sep=0pt, minimum width=0pt, minimum height=0pt, text width=0pt}
  ]
    \scriptsize
    \node (input) {$x$};

    \node (e1) [encoder,rotate=90, right= 4ex of input, anchor=north] {Encoder};
    \node (y1) [right= of e1.south, anchor=west] {$y$};
    \node (e2) [encoder,rotate=90, right=16ex of y1, anchor=north] {Hyper-Encoder};

    \node (z)     [right=of e2.south, anchor=center] {$z$};
    \node (quant) [coder, below=7ex of z.center, anchor=center] {Q};
    \node (ae)    [coder, below=7ex of quant.center, anchor=center] {AE};
    \node (ad)    [coder, below=7ex of ae.center, anchor=center] {AD};
    \node (zhat)  [below=7ex of ad.center, anchor=center] {$\hat{z}$};

    \node (quant1) [coder, below=7ex of y1.center, anchor=center] {Q};
    \node (ae1)    [coder, below=7ex of quant1.center, anchor=center] {AE};
    \node (ad1)    [coder, below=7ex of ae1.center, anchor=center] {AD};
    \node (yhat1)  [below=7ex of ad1.center, anchor=center] {$\hat{y_0}$};

    \node (entropy)  [entropy,right=8ex of ae1.east, anchor=center,rotate=90] {MoE-based Channel Group Entropy Model};

    \node (d1) [decoder,rotate=90, below=28ex of e1.center, anchor=center] {Decoder};
    \node (d2) [decoder,rotate=90, below=28ex of e2.center, anchor=center] {Hyper-Decoder};

    \node (output) [below=28ex of input.center, anchor=center] {$\hat{x}$};

    \draw[arrow] (e1) -- (y1);
    \draw[arrow] (y1) -- (e2);
    \draw[arrow] (e2) -- (z);
    \draw[arrow] (z) -- (quant);
    \draw[arrow] (quant) -- (ae);
    \draw[arrow] (ae) -- (ad);
    \draw[arrow] (ad) -- (zhat);
    \draw[arrow] (zhat) -- (d2);

    \draw[arrow] (y1) -- (quant1);
    \draw[arrow] (quant1) -- (ae1);
    \draw[arrow] (ae1) -- (ad1);
    \draw[arrow] (ad1) -- (yhat1);
    \draw[arrow] (yhat1) -- (d1);

    \draw[arrow] (d2.north) -| (entropy.west);
    \draw[arrow,dotted] (entropy) -- (ae1.east);
    \draw[arrow,dotted] (entropy) -- (ad1.east);

    \draw[arrow] (input) -- (e1);
    \draw[arrow] (d1) -- (output);

    \node (helper) [helper,below=8ex of d2.west] {};

  \end{tikzpicture}
    \caption{Overall architecture, following a similar structure to previous works.}\label{fig:arch_compression_model}
  \end{subfigure}
  \begin{subfigure}[t]{0.49\textwidth}
    \centering
    \tikzset{node distance = 4ex and 4ex}
\begin{tikzpicture}[
    module/.style={draw, thick, rounded corners, minimum width=15ex, minimum height=4ex,align=center, fill=gray!20},
    operation/.style={draw, minimum width=10ex, minimum height=3ex,align=center, fill=gray!20},
    embmodule/.style={module, fill=red!20},
    mhamodule/.style={module, fill=orange!20},
    lnmodule/.style={module, fill=yellow!20},
    ffnmodule/.style={module, fill=cyan!20},
    encoder/.style={module, fill=green!20, minimum width=15ex},
    decoder/.style={module, fill=green!20, minimum width=15ex},
    expert/.style={module, fill=blue!20, minimum width=15ex},
    arrow/.style={-stealth', thick},
    connector/.style={thick, rounded corners},
    coder/.style={module, minimum width=5ex, fill=yellow!20},
    helper/.style={inner sep=0pt, minimum width=0pt, minimum height=0pt, text width=0pt}
  ]
    \scriptsize

    \node (mean_scale) [] {$\hat z \rightarrow \mu, \sigma$};
    \node (concat) [circle, draw=black, right= 4ex of mean_scale] {C};
    \node (y_hat) [right=4ex of concat] {$\hat y \rightarrow \hat y_0 \ldots \hat y_{<i}$};

    \node (shortcut_split) [draw, shape = circle, fill = black, minimum size = 0.1cm, inner sep=0pt, below=4ex of concat] {};
    \node (norm) [below=2ex of shortcut_split,lnmodule] {Norm};
    
    \node(helper_in) [helper, below=4ex of norm]{};

    \node (proj_shared) [module, left=2ex of helper_in] {Conv};
    \node (exp_shared) [module, below=of proj_shared] {Self-Attention};

    \node (proj_experts) [module, right=2ex of helper_in] {Conv};
    \node (router) [module, below=of proj_experts] {Router};
    \node (experts_mult_3) [expert, below=of router,xshift=6pt,yshift=6pt] {};
    \node (experts_mult_2) [expert, below=of router,xshift=4pt,yshift=4pt] {};
    \node (experts_mult_1) [expert, below=of router,xshift=2pt,yshift=2pt] {};
    \node (experts) [expert, below=of router] {Expert};

    \node (merge) [lnmodule, below=of exp_shared] {Cross-Attention};

    \node (norm_out) [lnmodule, below=of merge] {FFN + Norm};
    \node (shortcut) [circle,draw, inner sep=0.5pt,below=2ex of norm_out] {$+$};
    \node (project) [module, below=2ex of shortcut] {Conv};

    \draw[arrow] (mean_scale) -- (concat);
    \draw[arrow] (y_hat) -- (concat);

    \draw[-,thick] (concat) -- (shortcut_split);
    \draw[arrow] (shortcut_split) -- ++(-20ex,0) |- (shortcut);
    \draw[-,thick] (shortcut_split) -- (norm);

    \draw[arrow] (norm.south) |- (proj_shared);
    \draw[arrow] (proj_shared) -- (exp_shared);
    \draw[arrow] (exp_shared) -- (merge);

    \draw[arrow] (norm.south) |- (proj_experts);
    \draw[arrow] (proj_experts) -- (router);
    \draw[arrow] (router) -- ([yshift=1ex]experts.north);
    \draw[arrow] (experts) -- (merge);

    \draw[arrow] (merge) -- (norm_out);
    \draw[-,thick] (norm_out) -- (shortcut);

    \draw[arrow] (shortcut) -- (project);

    \node (output_text) [right=of project,align=center] {$\mu,\sigma$-\textit{transforms}\\\textit{LRP}};
    \draw[arrow,dotted] (project) -- (output_text);

    \node (helper_left) [helper, left=4ex of proj_shared.west] {};
    \begin{scope}[on background layer]
      \node (group) [fit=(shortcut_split)(proj_experts)(proj_shared)(experts)(norm)(project)(experts_mult_3)(helper_left),dashed,rounded corners,inner sep=2ex,fill=blue!10,draw=blue!50,thick] {};
    \end{scope}

    \node (e_linear_in) [module, right=of group.north east,anchor=north west,yshift=-6ex] {Linear};
    \node (e_kqv_split) [draw, shape = circle, fill = black, minimum size = 0.1cm, inner sep=0pt, below=3ex of e_linear_in] {};

    \node (e_patchify) [operation, below=4ex of e_kqv_split,xshift=-2ex] {Patchify};
    \node (e_fft) [operation, below=3ex of e_patchify] {FFT};

    \node (e_attn) [circle,draw, inner sep=0.5pt,below=2ex of e_fft,xshift=2ex] {$\ast$};

    \node (e_ifft) [operation, below=2ex of e_attn] {IFFT};

    \node (e_v_mod) [circle,draw, inner sep=0.5pt,below=2ex of e_ifft] {$\ast$};

    \node (e_linear_out) [module, below=3ex of e_v_mod] {Linear};

    \draw[-,thick] (e_linear_in) -- (e_kqv_split);
    \draw[arrow] (e_kqv_split) -| (e_patchify.north);

    \draw[-,thick] ([xshift=2ex]e_patchify.south) -- ([xshift=2ex]e_fft.north);
    \draw[-,thick] ([xshift=-2ex]e_patchify.south) -- ([xshift=-2ex]e_fft.north);

    \draw[arrow] ([xshift=-2ex]e_fft.south) |- (e_attn.west);
    \draw[-,thick] ([xshift=2ex]e_fft.south) -- (e_attn.north);
    \draw[-,thick] (e_attn) -- (e_ifft);
    
    \draw[-,thick] (e_ifft) -- (e_v_mod);

    \draw[-,thick] (e_kqv_split) -- ++(6ex,0) |- (e_v_mod.east);
    \draw[arrow] (e_v_mod) -- (e_linear_out);

    \node (e_group) [fit=(e_linear_in)(e_linear_out),draw,dotted,label=above:Expert] {};
    \draw[dotted] ([yshift=2ex]experts.north east) -- (e_group.north west);
    \draw[dotted] (experts.south east) -- (e_group.south west);

  \end{tikzpicture}
    \caption{MoE-based Channel Group Entropy Model}\label{fig:arch_moe_entropy}
  \end{subfigure}
  \caption{Architecture overview. We use a channel-group entropy model, which is commonly used in other learned image compression methods. A Mixture of Experts model is used to perform context prediction using the already decoded hyper-prior $\hat z$ and slices $y_{<i}$ to provide the input for mean and scale estimation.}\label{fig:architecture}
\end{figure*}
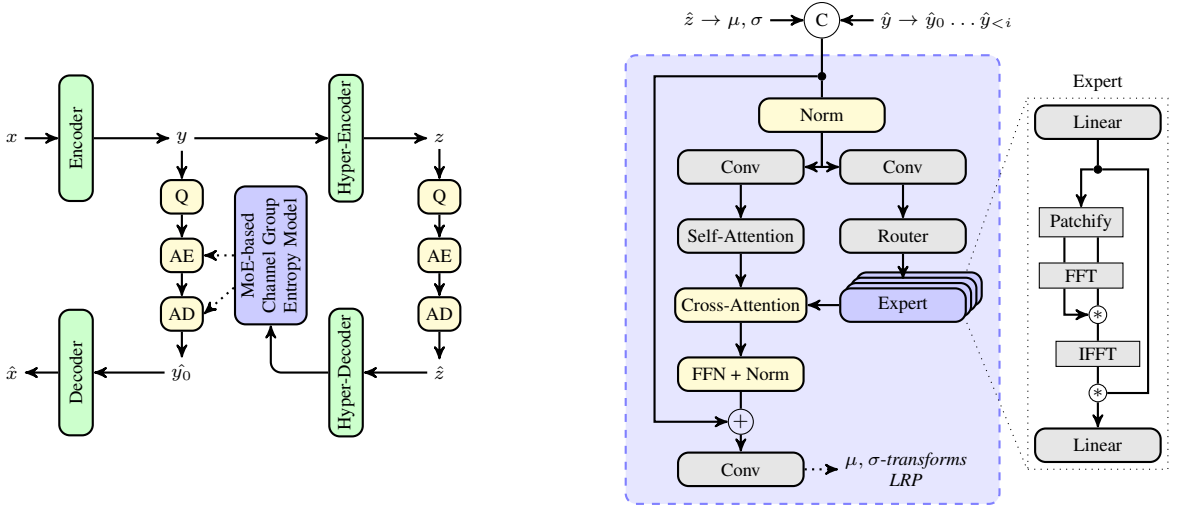

We explore the use of Mixture of Experts (MoE) layers in learned image compression (LIC). 
Specifically, we propose a MoE-based entropy model to improve the compression efficiency of learned image compression models.

\subsection{MoE-based Entropy Model}\label{sec:method:moee}
DCAE~\cite{lu2025learned} introduced a dictionary learning approach to adapt the entropy model to different image types. 
Motivated by the same goal, we propose to use MoE layers instead of the dictionary learning approach. 

As shown in Fig.~\ref{fig:arch_compression_model}, we follow the same common structure for learned compression models as other recent works.
The analysis and synthesis transforms of the compression model are based on the DCAE~\cite{lu2025learned} model, using the same Swin-Transformer~\cite{liu2021swin} layers.
The entropy model is based on the grouped-channel entropy model~\cite{he2022elic,jiang2023mlic,lu2025learned,feng2025linear}.

Specifically, the latent $y$ is split into multiple channel groups $y_i$ (in our case we use 5 groups), where each group is modeled sequentially using the hyperprior $\hat z$ and any previously decoded channel groups $\hat y_{<i}$ as context. After quantization of a slice, a latent residual prediction module refines the quantized latent representation after each slice is decoded.

Our MoE-based context model, shown in Fig.~\ref{fig:arch_moe_entropy}, consists of a gating network and several expert networks. 
The gating network is realized as a linear layer for per sample routing, similar to previous works~\cite{zamfir2025complexity}.
For every image in a batch, an expert is selected according to the gating scores of each expert. 
In addition to the experts, a shared branch allows for a more efficient use of resources and is realized as a simple self-attention layer. 

During training, we use noisy softmax top-1 routing~\cite{zamfir2025complexity}, with $\epsilon \sim \mathcal{N}\left(0, \frac{1}{n^2}\right)$ where $n$ is the number of experts, as defined in Eq.~\ref{eq:gating}.
During the actual compression, the selection needs to be fully deterministic, i.e. $\epsilon = 0$.
\begin{equation}\label{eq:gating}
  g(x) = top_k\left(\text{Softmax}\left(\text{W}x + \epsilon\right)\right)
\end{equation}

The expert architecture follows~\cite{zamfir2025complexity}, and is realized as a feedforward neural network using FFT-based attention layers~\cite{kong2023efficient} for an efficient approximation of the attention mechanism.  

The different experts use increasing channel dimensions to reduce the computational complexity.
This means for any collection of experts for some slice $y_i$ with input dimension $k_i=2M + \frac{M\cdot i}{5} $ the individual experts $e_{j \in \left[0, 3\right]}$ use channel dimensions of $\lfloor \frac{k_i}{2^{j}} \rfloor$.
While using the same channel dimensions over all experts is possible, we did not find it to be helpful (See Sec.~\ref{sec:ablations}).

Balanced selection of experts is achieved using an additional auxiliary loss $\mathcal{L}_{aux}$ (see Eq.~\ref{eq:aux_loss}) that is based on the selection of experts across different samples in a batch~\cite{zamfir2025complexity,riquelme2021scaling}.
This auxiliary loss consists of two components: importance and load.
The importance loss is based on the gating scores of each expert across the batch.
\begin{equation}\label{eq:aux_loss}
    \mathcal{L}_{aux} = 0.5 \cdot CV(Imp)^2 + 0.5 \cdot CV(Load)^2
\end{equation}
Unlike~\cite{zamfir2025complexity}, we do not account for expert complexity in the importance loss. %

\begin{figure*}[t]
    \centering
    \scriptsize
    \begin{subfigure}[t]{0.33\textwidth}
      \centering
      \begin{tikzpicture}
    \scriptsize
    \begin{axis}[
        yticklabel style={
            /pgf/number format/precision=2
        },
        xticklabel style={
            /pgf/number format/fixed,
            /pgf/number format/precision=3
        },
        scaled x ticks=false,
        xlabel=bits-per-pixel (BPP) \textdownarrow,
        ylabel style = {align=center},
        ylabel=PSNR \textuparrow,
        ylabel near ticks,
        width=\textwidth-4em,
        height=8cm,
        mark size=1.5pt,
        grid=both,
        grid style={dashdotted},
        minor grid style={dotted,gray!30},
        minor tick num=3,
        legend cell align=left,
        legend pos=south east,
        scale only axis,
        xmin=0.08,
        xmax=0.9
    ]

        \addplot [moe] coordinates {  %
            (0.11599392350763083, 29.432170073191326)
            (0.181, 30.903)
            (0.272, 32.433)
            (0.412, 34.352)
            (0.586, 36.189)
            (0.7972,37.9271)
        }; \label{psnr_kodak_moe}

        \addplot [dcae] coordinates {  %
            (0.110,29.243)
            (0.176,30.736)
            (0.293,32.677)
            (0.428,34.407)
            (0.595,36.109)
            (0.813,37.869)
        }; \label{psnr_kodak_dcae}

        \addplot [lalic] coordinates {  %
            (0.115,29.253)
            (0.186,30.855)
            (0.286,32.556)
            (0.426,34.312)
            (0.601,36.093)
            (0.831,37.935)
        }; \label{psnr_kodak_lalic}

        \addplot [cca] coordinates {  %
           (0.400,33.915)
           (0.515,35.189)
           (0.631,36.282)
           (0.744,37.198)
        }; \label{psnr_kodak_cca}

        \addplot [mlicpp] coordinates {  %
            (0.107,29.179)
            (0.174,30.744)
            (0.272,32.379)
            (0.412,34.235)
            (0.592,35.912)
            (0.8020,37.4594)
        }; \label{psnr_kodak_mlicpp}

        \addplot [tcm] coordinates {  %
            (0.162,30.042)
            (0.204,30.856)
            (0.312,32.531)
            (0.455,34.136)
            (0.653,36.056)
            (0.899,37.979)
        }; \label{psnr_kodak_tcm}

        \addplot [elic] coordinates {  %
            (0.044, 26.105)
            (0.074, 27.490)
            (0.122, 28.955)
            (0.195, 30.526)
            (0.497, 34.490)
            (0.865, 37.527)
        }; \label{psnr_kodak_elic}
        
        \addplot [vvc] coordinates {  %
            (0.1123589409722222,28.43264744879213)
            (0.155197991265191,29.45954512799564)
            (0.2119861178927952,30.53104382897368)
            (0.286176045735677,31.683147282597982)
            (0.3765860663519966,32.82541900282706)
            (0.4905403984917535,34.01671650725221)
            (0.6265335083007811,35.18614069393125)
            (0.7841169569227432,36.33389517632819)
            (0.971261766221788,37.43298582654395)
        }; \label{psnr_kodak_vvc}

        \legend{
            MoEE (ours),
            DCAE (CVPR'25),
            LALIC (CVPR'25),
            CCA (NeurIPS'24),
            MLIC++ (ICMLW'23),
            TCM (CVPR'23),
            ELIC (CVPR'22),
            VVC (VTM-17.0), 
        }
    \end{axis}
\end{tikzpicture}
      \caption{Kodak images~\cite{kodak1993kodak}}\label{fig:rd_kodak}
    \end{subfigure}
    \begin{subfigure}[t]{0.33\textwidth}
      \centering
      \begin{tikzpicture}
    \scriptsize
    \begin{axis}[
        yticklabel style={
            /pgf/number format/precision=2
        },
        xticklabel style={
            /pgf/number format/fixed,
            /pgf/number format/precision=3
        },
        scaled x ticks=false,
        xlabel=bits-per-pixel (BPP) \textdownarrow,
        ylabel style = {align=center},
        ylabel=PSNR \textuparrow,
        ylabel near ticks,
        width=\textwidth-4em,
        height=8cm,
        mark size=1.5pt,
        grid=both,
        grid style={dashdotted},
        minor grid style={dotted,gray!30},
        minor tick num=3,
        legend cell align=left,
        legend pos=south east,
        scale only axis,
        xmin=0.08,
        xmax=0.65,
    ]        
        \addplot [moe] coordinates {
            (0.0980993334017694, 31.5917129325867) %
            (0.1405015552,32.99766469)
            (0.1961777783, 34.3293962)
            (0.2836031103, 35.85846741)
            (0.3938624435, 37.32624727)
            (0.5384246689, 38.69267227)
        }; \label{psnr_tecnick_moe}

        \addplot [dcae] coordinates {  %
            (0.094,31.430)
            (0.136,32.832)
            (0.209,34.475)
            (0.292,35.903)
            (0.400,37.255)
            (0.552,38.646)
        }; \label{psnr_kodak_dcae}

        \addplot [lalic] coordinates {  %
            (0.099,31.357)
            (0.146,32.866)
            (0.203,34.341)
            (0.291,35.822)
            (0.405,37.259)
            (0.564,38.742)
        }; \label{psnr_kodak_lalic}
        
        \addplot [cca] coordinates {  %
            (0.278, 35.439)
            (0.352, 36.482)
            (0.429, 37.365)
            (0.506, 38.107)
        }; \label{psnr_kodak_cca}

        \addplot [mlicpp] coordinates {  %
            (0.090,31.284)
            (0.135,32.743)
            (0.195,34.140)
            (0.289,35.692)
            (0.403,37.016)
            (0.553,38.331)
        }; \label{psnr_kodak_mlicpp}

        \addplot [tcm] coordinates {  %
            (0.130,32.021)
            (0.157,32.788)
            (0.225,34.246)
            (0.315,35.570)
            (0.446,37.100)
            (0.622,38.650)
        }; \label{psnr_kodak_tcm}

        \addplot [elic] coordinates {  %
            (0.034, 27.937)
            (0.056, 29.363)
            (0.091, 30.814)
            (0.143, 32.296)
            (0.353, 35.742)
            (0.627, 38.271)
        }; \label{psnr_kodak_elic}

        \addplot [vvc] coordinates {  %
            (0.0917141111111111,30.187271359488488)
            (0.1214761666666666,31.19434031468773)
            (0.1595285555555555,32.206911844427005)
            (0.2075158333333333,33.28703546845274)
            (0.266551388888889,34.25914654249331)
            (0.3407139444444444,35.22066358622896)
            (0.4319719444444445,36.14668223341316)
            (0.5448062777777778,37.04811997240603)
            (0.6866582222222222,37.93522785041841)
        }; \label{psnr_tecnick_vvc}

        \legend{
            MoEE (ours),
            DCAE (CVPR'25),
            LALIC (CVPR'25),
            CCA (NeurIPS'24),
            MLIC++ (ICMLW'23),
            TCM (CVPR'23),
            ELIC (CVPR'22),
            VVC (VTM-17.0), 
        }
    \end{axis}
\end{tikzpicture}
    \caption{Tecnick test images~\cite{asuni2014testimages}}\label{fig:rd_tecnick}
    \end{subfigure}
    \begin{subfigure}[t]{0.33\textwidth}
      \centering
      \begin{tikzpicture}
    \scriptsize
    \begin{axis}[
        yticklabel style={
            /pgf/number format/precision=2
        },
        xticklabel style={
            /pgf/number format/fixed,
            /pgf/number format/precision=3
        },
        scaled x ticks=false,
        xlabel=bits-per-pixel (BPP) \textdownarrow,
        ylabel style = {align=center},
        ylabel=PSNR \textuparrow,
        ylabel near ticks,
        width=\textwidth-4em,
        height=8cm,
        mark size=1.5pt,
        grid=both,
        grid style={dashdotted},
        minor grid style={dotted,gray!30},
        minor tick num=3,
        legend cell align=left,
        legend pos=south east,
        scale only axis,
        xmin=0.05,
        xmax=0.65
    ]       

        \addplot [moe] coordinates {
            (0.0883887462863108, 31.346516120724562)
            (0.135534266089754, 32.7192779858907)
           (0.1976418896782689, 34.091273842788326)
           (0.2955471789691506, 35.68018396889291)
           (0.4204452108682656,37.217118891274055)
           (0.5810226500034332, 38.65811036272747)
        }; \label{psnr_clic_moe}

        \addplot [dcae] coordinates {  %
            (0.084,31.181)
            (0.131,32.575)
            (0.213,34.245)
            (0.307,35.712)
            (0.428,37.116)
            (0.596,38.580)
        }; \label{psnr_clic_dcae}

        \addplot [lalic] coordinates {  %
            (0.089,31.164)
            (0.141,32.651)
            (0.207,34.149)
            (0.304,35.664)
            (0.431,37.157)
            (0.607,38.708)
        }; \label{psnr_clic_lalic}

        \addplot [mlicpp] coordinates {  %
            (0.080,31.089)
            (0.129,32.541)
            (0.196,33.971)
            (0.305,35.545)
            (0.429,36.931)
            (0.591,38.247)
        }; \label{psnr_clic_mlicpp}

        \addplot [cca] coordinates {  %
            (0.286, 35.251)
            (0.368, 36.317)
            (0.453, 37.238)
            (0.538, 38.011)
        }; \label{psnr_clic_cca}

        \addplot [tcm] coordinates {  %
            (0.120,31.884)
            (0.150,32.633)
            (0.225,34.105)
            (0.323,35.452)
            (0.467,37.047)
            (0.660,38.678)
        }; \label{psnr_clic_tcm}
        
        \addplot [elic] coordinates {  %
            (0.034, 27.937)
            (0.056, 29.363)
            (0.091, 30.814)
            (0.143, 32.296)
            (0.353, 35.742)
            (0.627, 38.271)
        }; \label{psnr_clic_elic}

        \addplot [vvc] coordinates { 
            (0.0821876974257416,30.21450349118233)
            (0.1130866860723925,31.180124982936093)
            (0.1539918833026654,32.16754964142189)
            (0.2058784495216643,33.191267391414456)
            (0.2701131902158922,34.16623377986386)
            (0.3507965112097417,35.13625455539849)
            (0.4497492904131193,36.09042552293772)
            (0.5707965149537835,37.02000022140685)
            (0.7220543632106404,37.9402084346042)
        }; \label{psnr_clic_vvc}

        \legend{
            MoEE (ours),
            DCAE (CVPR'25),
            LALIC (CVPR'25),
            MLIC++ (ICMLW'23),
            CCA (NeurIPS'24),
            TCM (CVPR'23),
            ELIC (CVPR'22),
            VVC (VTM-17.0), 
        }
    \end{axis}
\end{tikzpicture}
    \caption{CLIC 2020 Professional~\cite{toderici2020workshop}}\label{fig:rd_clic}
    \end{subfigure}
    \caption{Rate-distortion performance on various datasets.}\label{fig:rd_plot}
\end{figure*}
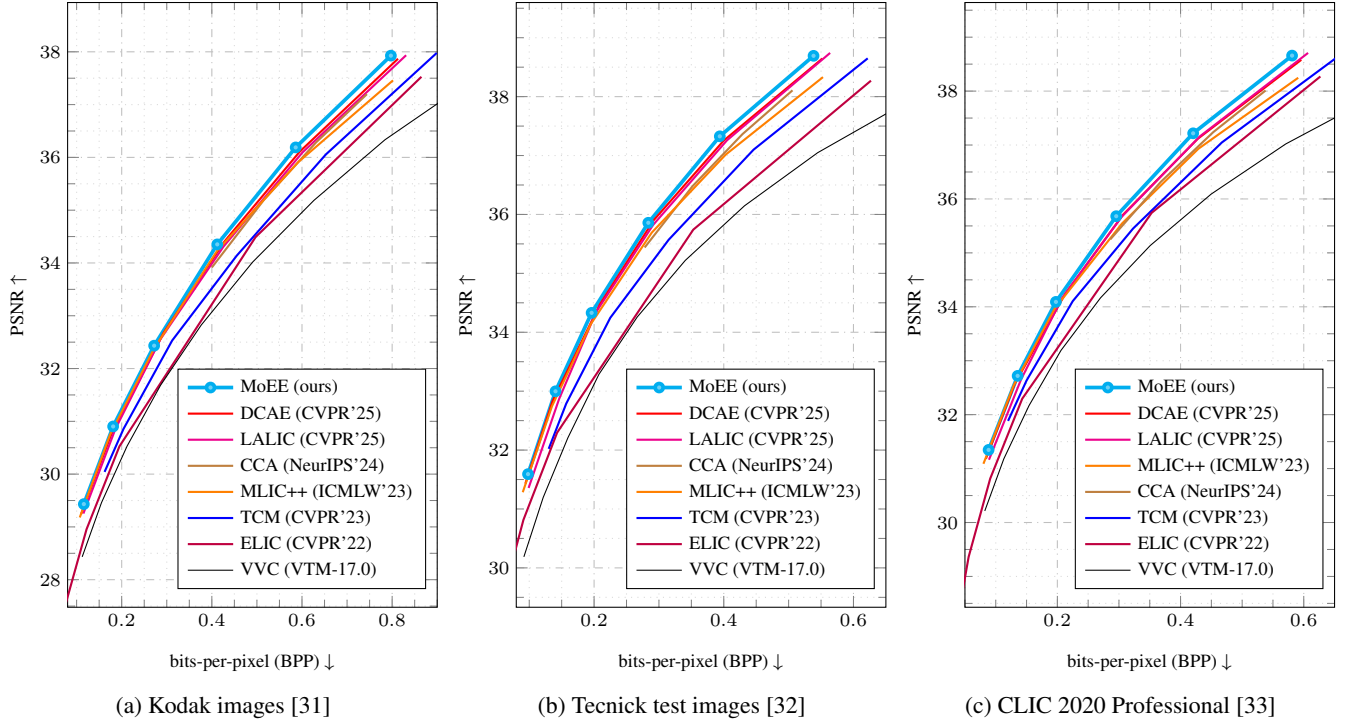
\begin{table*}[t]
  \centering
  \begin{tabular}{lccccc} %
    \toprule
    Model & BD-Rate \textdownarrow & Params \textdownarrow & max. VRAM \textdownarrow & dec. latency (ms) \textdownarrow \\
    \midrule
    MoEE (ours) & \textbf{-16.85\%} & 175M & 1.73GB & 73 \\
    DCAE (CVPR'25)~\cite{lu2025learned} & -15.06\% & 119M & 1.49GB & 67 \\
    LALIC (CVPR'25)~\cite{feng2025linear} & -13.74\% & 66M & 0.95GB & 97 \\
    MLIC++ (ICMLW'23)~\cite{jiang2023mlic++} & -14.69\% & 117M & 1.42GB & 98 \\
    CCA (NeurIPS'24)~\cite{han2024causal} & -10.89\% & 65M & 0.91GB  & 83 \\
    TCM (CVPR'23)~\cite{liu2023learned} & -5.43\% & 77M & 1.83GB  & 93 \\
    ELIC (CVPR'22)~\cite{he2022elic} & -2.40\% & \textbf{34M} & \textbf{0.43GB}  & \textbf{61} \\
    \bottomrule
  \end{tabular}
  \caption{Quantitative performance comparison of recent methods, evaluated on 
  the Kodak images~\cite{kodak1993kodak}. VVC (VTM-17.0) serves as an anchor for the BD-Rate calculation.
  Results are averaged over 5 runs on an NVIDIA RTX 4090 GPU.}\label{tab:compute}
\end{table*}

\subsection{Loss function}
We follow the typical rate-distortion optimization approach for learned image compression~\cite{balle2018variational}.
The complete loss function for training to optimize the rate-distortion trade-off is given by:
\begin{equation}\label{eq:loss}
    \mathcal{L} = R + \lambda D + 0.01 \cdot \mathcal{L}_{aux}
\end{equation}

Where $R$ is the estimated bitrate, $D$ is the distortion (measured in MSE) and $\lambda$ is the Lagrangian multiplier which controls the rate-distortion trade-off.

Similarly to previous works~\cite{he2022elic,jiang2023mlic}, quantization during training is approximated using additive uniform noise when computing the rate term. When computing the distortion term, we use straight-through estimation~\cite{theis2017lossy}.

\subsection{SGA latent refinement}
Yang et al.~\cite{yang2020improving} proposed stochastic Gumbel annealing (SGA) latent refinement to improve the quality of the latent representation by refining it before transmission. 
We apply this refinement to further improve the latent representation, at the cost of slower encoding.

During the refinement process, an optimizer is used to refine the latent representation using the same RD-loss and $\lambda$ as the original model (without $\mathcal{L}_{aux}$ which only affects routing across multiple images), by backward propagating the gradients through the decoder. 
Using SGA to quantize the latent representation allows us to refine the latent representation to correct errors introduced due to the approximation of the quantization operation during training.

For our experiments, we optimize for 3000 SGA iterations with an annealing rate of $10 ^{-3}$ and a learning rate of $10 ^{-4}$ with the Adam~\cite{kingma2014adam} optimizer.

\section{Experiments}
\label{sec:experiments}

\begin{table}[t]
  \centering
  \begin{tabular}{lccc}
    \toprule
    Variant & dec. (ms) & Params & BD-Rate \\
    \midrule
    Constant dim. & 73 & 227M & +0.03\% \\   %
    Spread dim. (default) & 73 & \textbf{175M} & \textbf{0.00\%} \\
    \bottomrule
  \end{tabular}
  \caption{RD-performance of using constant channel dimension across all experts vs increasing channel dimensions, evaluated on the Kodak~\cite{kodak1993kodak} images.}\label{tab:abl:rank}
\end{table} %
\subsection{Experimental Setup}

We train our model on the same dataset as MLIC~\cite{jiang2023mlic}, which consists of 100k images at a resolution of 512\texttimes 512 pixels. The images in this dataset are selected from ImageNet~\cite{deng2009imagenet}, COCO 2017~\cite{lin2014microsoft}, DIV2K~\cite{agustsson2017ntire} and Flickr2K~\cite{lim2017enhanced}.

Following previous work, we train for 2M steps using the Adam optimizer~\cite{kingma2014adam} with a learning rate of $10^{-4}$, and a batch size of 16. 
We use $\lambda=0.0130$ for the base model.
Afterwards, we finetune with the final $\lambda$-value for an additional 500k steps using a smaller learning rate of $10^{-5}$ to obtain checkpoints for different bitrates. %
Finally, we finetune for an additional 200k steps using higher resolution crops of 512\texttimes 512 pixels.

The source code is available on our project page\footnote{Project page: https://jbrenig.github.io/ICIP26-MoEE}.

\subsection{Rate-Distortion performance}
We evaluate our model on standard benchmark datasets Kodak~\cite{kodak1993kodak}, Tecnick~\cite{asuni2014testimages} and CLIC 2020 Professional~\cite{toderici2020workshop}. 

The Kodak dataset consists of 24 images with a resolution of 768\texttimes 512 pixels. The Tecnick dataset contains 100 images with a resolution of 1200\texttimes 1200 pixels. The CLIC 2020 validation set consists of 41 images with varying resolutions around 2048\texttimes 1440 pixels.

In Fig.~\ref{fig:rd_plot} we present our results in terms of PSNR and bits-per-pixel. 
For most bitrates, our model achieves better rate-distortion performance compared to the state-of-the-art methods on all datasets. The improvement is most pronounced at higher bitrates, while for very low bitrates our model performs on par with existing methods. 
In terms of BD-Rate, we achieve a $-16.85\%$ improvement over the VVC (VTM-17.0)~\cite{dominguez2022versatile} baseline on the Kodak images. 
For the Tecnick~\cite{asuni2014testimages} and  CLIC~\cite{toderici2020workshop} datasets, BD-Rate improvement are $-22.09\%$ and $-8.08\%$, respectively.

\begin{figure*}[t]
    \footnotesize
    \centering
    \input{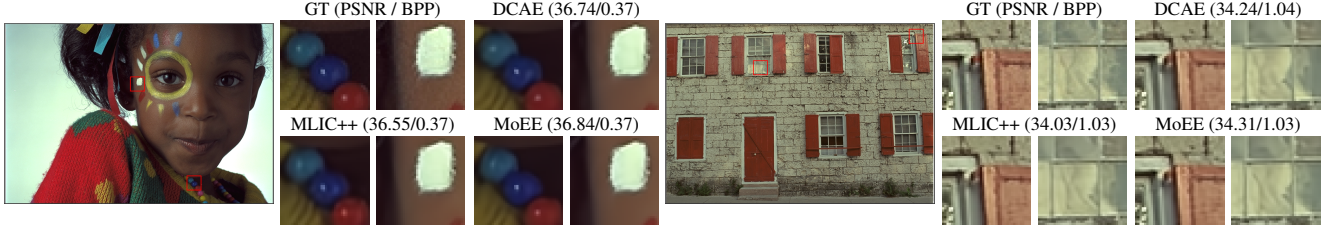}
    \caption{Visual comparison of our MoEE approach with the recent DCAE~\cite{lu2025learned} and MLIC++~\cite{jiang2023mlic++} methods.}
    \label{fig:img:compare}
\end{figure*}
\subsection{Ablation Study with increased expert capacity}\label{sec:ablations}

As described in Sec.~\ref{sec:method:moee}, we use the same spread channel configuration as in \cite{zamfir2025complexity}, which means that the different experts are of different sizes. 
Using a fixed (higher) size for all experts increases the number of parameters. 
To evaluate the impact of this decision, we train the networks for 1M steps and finetune to the final bitrate for another 50k steps at a lower learning rate.
As seen in Tab.~\ref{tab:abl:rank}, the increased capacity of the experts does not lead to better performance.

\begin{table}[t]%
  \centering
  \begin{tabular}{lccc}
    \toprule
    Variant & dec (ms.) & Params & BD-Rate \\ 
    \midrule
    no MoE              & \textbf{73} & \textbf{163M} & +1.73\% \\ 
    2 Experts           & \textbf{73} & 171M          & +1.56\% \\  
    4 Experts (default) & \textbf{73} & 175M          & \textbf{0.00\%} \\ 
    6 Experts           & 74          & 176M          & +0.59\% \\ 
    \bottomrule
  \end{tabular}
  \caption{RD-performance for a different number of experts, evaluated on the Kodak~\cite{kodak1993kodak} images.}\label{tab:abl:num_experts} %
\end{table}

\subsection{Ablation Study with different number of experts}
For the main experiments, the model uses 4 experts. In Tab.~\ref{tab:abl:num_experts} we also evaluate the performance with different numbers of experts. For this ablation, we train the networks for 400k steps.
We found 4 experts to be a good trade-off between performance and computational complexity. %

\subsection{Computational cost}
Mixture of Experts architectures are known to be computationally efficient, as only a subset of experts is activated for each input.
This means that the model can be scaled up in terms of the number of parameters without a significant increase in computational cost.
Overall, the computational cost of our MoE-based model is comparable to other state-of-the-art models while achieving better rate-distortion performance.

As seen in Tab.~\ref{tab:compute}, even though our model uses more parameters than other recent models, the decoding latency remains comparable, since only one expert is activated per input image.

\section{Conclusion}
\label{sec:conclusion}

In this paper, we proposed a Mixture of Experts-based entropy model for learned image compression. 
Our approach allows the model to select specialized experts based on the input image while activating only a subset of the model's parameters, leading to improved compression performance. 
Extensive experiments demonstrated that our MoE-based LIC model outperforms state-of-the-art learned and traditional image compression methods on standard benchmarks. 
Future work includes exploring more advanced MoE architectures and further optimizing the computational efficiency of the model.

{
\small
\bibliographystyle{IEEEbib}
\bibliography{refs}
}

\end{document}